\documentclass{article}
\usepackage{spconf,amsmath,graphicx,hyperref}

\usepackage{cite}
\usepackage{multirow}
\usepackage{xspace} 
\usepackage{algorithm,algpseudocode}
\usepackage{lipsum}
\usepackage{subcaption}
\usepackage{multirow}
\usepackage{makecell}
\usepackage{float}
\usepackage{booktabs}
\usepackage{comment}

\title{Unsupervised Video Class-Incremental Learning \\ via Deep Embedded Clustering Management}
\name{Nattapong Kurpukdee, Adrian G. Bors }
\address{Department of Computer Science,\\
University of York, York YO10 5GH, UK\\
    nattapong.kurpukdee@york.ac.uk, adrian.bors@york.ac.uk}
\begin{document}
%
\maketitle
\begin{abstract}
Unsupervised video class incremental learning (uVCIL) represents an important learning paradigm for learning video information without forgetting, and without considering any data labels. Prior approaches have focused on supervised class-incremental learning, relying on using the knowledge of labels and task boundaries, which is costly, requires human annotation, or is simply not a realistic option. In this paper, we propose a simple yet effective approach to address the uVCIL. We first consider a deep feature extractor network, providing a set of representative video features during each task without assuming any class or task information. We then progressively build a series of deep clusters from the extracted features. During the successive task learning, the model updated from the previous task is used as an initial state in order to transfer knowledge to the current learning task. We perform in-depth evaluations on three standard video action recognition datasets, including UCF101, HMDB51, and Something-to-Something V2, by ignoring the labels from the supervised setting. Our approach significantly outperforms other baselines on all datasets.
\end{abstract}

\section{Introduction}
\label{sec:intro}

Continual learning is essential in video learning applications where information is continually acquired and requires processing. Applications such as environmental monitoring, robot behavior, health tracking, and CCTV surveillance all benefit from continual video learning. However, when retraining a system, previously trained on a video task, it forgets the information learnt previously. This is due to the fact that the system's parameters, estimated during the learning of the previous task, are replaced by new values when learning a new task. The collapse in performance in such cases is due to the catastrophic forgetting \cite{LifeLong_review}. A typical unsupervised video class-incremental. Training data emerge sequentially in the stream format (task) with new training dataset, which without any annotations, would require processing and analysis. 

Moreover, most existing video class-incremental learning are focusing on supervised scenarios are adaptations of methods initially developed for image such as the iCaRL \cite{iCaRL}, and BiC \cite{BIC}, have been extended to video domain in \cite{CIL-FOR-ACTION-CLASSIFICATION,CIL-FOR-ACTION-RECOGNITION-IN-VIDEOS,vCLIMB}. These models employ memory buffers to store videos, aiming to address catastrophic forgetting. Furthermore, the proposed for video such as \cite{CIL-FOR-ACTION-CLASSIFICATION, maraghi2022class, CIL-FOR-ACTION-RECOGNITION-IN-VIDEOS, pei2023space, vCLIMB}, also use memory buffers or prompts to retain the knowledge of prior classes with a CNN backbone as feature extractor. The temporal transformer features has provided a promising results in \cite{10647854}. Recently, the integration of large language models (LLMs) and vision models for video class-incremental \cite{pei2023space, PIVOT_villa} has been shown to be a promising area of research.

However, in supervised video class-incremental methods are still limited in their real-world applicability due to their reliance on labels, obtained with the help of costly human annotation, which remains a significant limitation in real-life applications. To address this challenges, we propose an approach for unsupervised video class-incremental learning. The proposed methodology ensures the continuous acquisition of information while optimizing both memory consumption and computational efficiency, without considering any data labels. We propose an approach entitled Unsupervised Video Class-Incremental Learning (uVCIL), which enables deep models to continually learn from unlabeled video streams without forgetting previously acquired knowledge. 

Our main contributions are as follows: (1) We propose an effective and practical approach which addresses the unexplored area of uVCIL by progressively building and managing a system of clusters consisting of representative features; (2) We address the cluster imbalance problem in the unsupervised video class-incremental learning, which is a major challenge when video data for certain clusters is not available in sufficient numbers; (3) We incorporate knowledge-preserving techniques to enhance learning retention, while also enabling the learning of new information ensuring stability-plasticity balance for the unsupervised video class-incremental learning settings; (4) We establish a benchmark evaluation protocol for uVCIL, providing extensive experimental analyses results to assess the effectiveness of our approach.

\begin{figure*}[ht!]
  \centering
  \includegraphics[width=\linewidth]{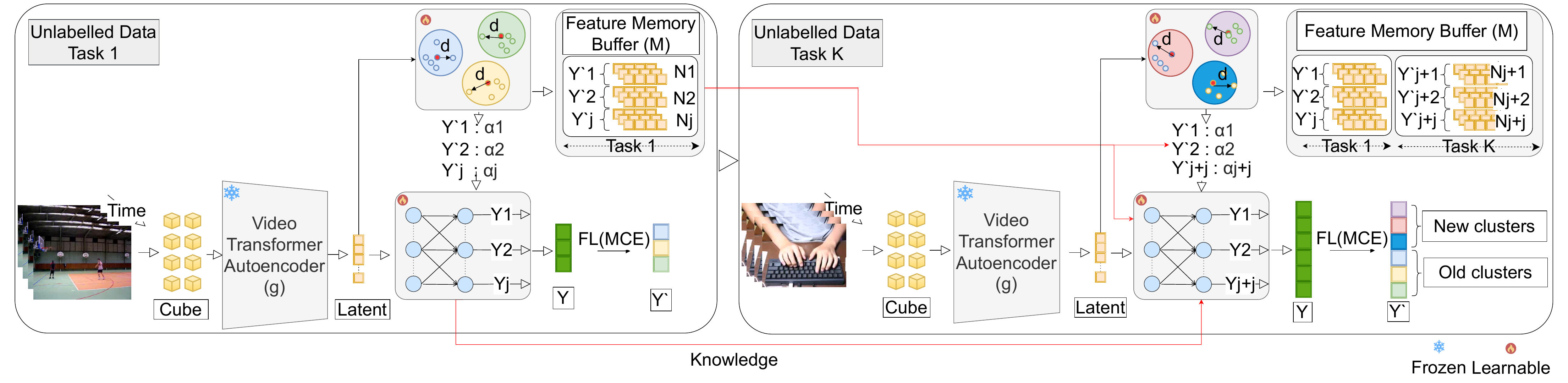}
  \vspace{-0.7cm}
  \caption{Overview of uVCIL. A video dataset will be divided into a series of tasks $\{ \tau_1,\tau_2,\ldots,\tau_k \}$. Each task data will feeds into a deep feature extractor $g(\cdot)$, which is used to extract the video features. After that, each task $\tau_k$, a fixed number of deep embedded clusters is formed, and each cluster has a memory buffer associated in order to store $N$ feature vector for that cluster. For every new task, the feature from the memory buffers is reused through memory replay to tackle forgetting. Moreover, novel data will form new clusters, contributing to the augmentation of the stored information.}
  \label{fig:proposed}
  \vspace{-0.5cm}
\end{figure*}

\vspace{-0.3cm}
\section{uVCIL methodology}
\label{sec:method}

\vspace{-0.3cm}
\subsection{Problem Setup}
\label{seg:problem_setup}
\vspace{-0.2cm}
We follow the supervised video class-incremental learning vCLIMB benchmark \cite{vCLIMB} where we ignoring the label for unsupervised video class-incremental learning. Our objective is to learn a pseudo-labeling function $f(\cdot)$, parameterized by a deep learning system, that assigns pseudo-labels to the data, $\tilde{y}_i = f({\bf v}_i)$, where we assume that the ideal label $y_i$ is unknown to $f(\cdot)$, where $i$ is the label's identifier. Then the model will learn to predict pseudo-labels $\tilde{y}_i$ given an input video $({\bf v}_i, \tilde{y}_i)$ from a sequence of tasks $\{\tau_1, \tau_2, \dots, \tau_k\}$, where ${\bf v}_i \subseteq X$ and $\tilde{y}_i \subseteq \tilde{Y}$. In the class-incremental, we setup the different tasks contain the data with a different set of labels without overlap with the other task data, meanwhile we ignore the class boundary for unsupervised learning. We define $CAcc_k$ as the cluster accuracy of tasks $k$. Our primary objective is to train a single model that maximizes the average cluster accuracy ACAcc, which is the average over all observed tasks include the final task $k$.

\vspace{-0.3cm}
\subsection{Assigning Pseudo-labels through deep clustering}
\label{seg:clustering_to_label}
\vspace{-0.2cm}

Firstly, we assume that we have extracted a set of features $\{{\bf x}_{k,i} = g ({\bf v}_i)\}_{i=1}^{n_k}$, using a pre-trained encoder network. The encoder network $g( \cdot)$ is used to extract the features when learning all tasks $\{ \tau_k | k=1,\ldots,K\}$, without being retrained, thus ensuring a consistent feature space over the entire data space. We consider deep embedded clustering by using $k$-means, which requires a pre-defined number of clusters $k$ in advance, which was used for video retrieval \cite{K-MeanVideo} and was adapted through deep embedding clustering for being used in the context of deep learning \cite{DeepClustering}. In the following, we apply a deep embedded clustering algorithm for the class-incremental learning of unsupervised video and call this method uVCIL-Clustering (uVCIL-CLU). This approach does not require any memory buffer. For each task $\tau_k$, we consider a fixed number of clusters, let us consider that as $l$. We first calculate the distances between the features for each video and the existing cluster centers as a measure of similarity with a specific grouping of data~:
\vspace{-0.2cm}
\begin{equation}
 d_j ({\bf v}_{j,i}) = \parallel g({\bf v}_{i} ) - \mbox{\boldmath$\mu$}_{j} \parallel .
 \label{AssignCluster} 
 \vspace{-0.2cm}
\end{equation}
We then produce pseudo-label cluster assignments $\tilde{y}_{i}$, for each data ${\bf v}_{i}$~:
\vspace{-0.5cm}
\begin{equation}
   \tilde{y}_{i} =  \arg\min_{j=1}^{l_k} d_j ({\bf v}_{j,i}) \} \ ,
  \label{eq:cluster_assignment}
  \vspace{-0.2cm}
\end{equation}
where ${\bf v}_{j,i}$ is closest to the cluster defined by the center $\mbox{\boldmath$\mu$}_m$ and its pseudo-label $\tilde{y}_i=m$ represents the assignment of the $i$'th sample to the cluster $m$, which represents an association with its cluster center $\mbox{\boldmath$\mu$}_m$, and $l_k$ is the number of clusters decided when learning the task ${\tau_k}$. The cluster centers $\mbox{\boldmath$\mu$}_j$, $j=1,\ldots,L_k$  created during the learning of the earlier tasks are frozen and reused when learning the subsequently learnt tasks, thus maximizing the forward learning transfer while minimizing both memory requirements and computation costs.

\begin{figure*}[t!]
     \centering
     \subfloat[UCF101 Cluster Accuracy]{
            \includegraphics[width=0.32\textwidth]{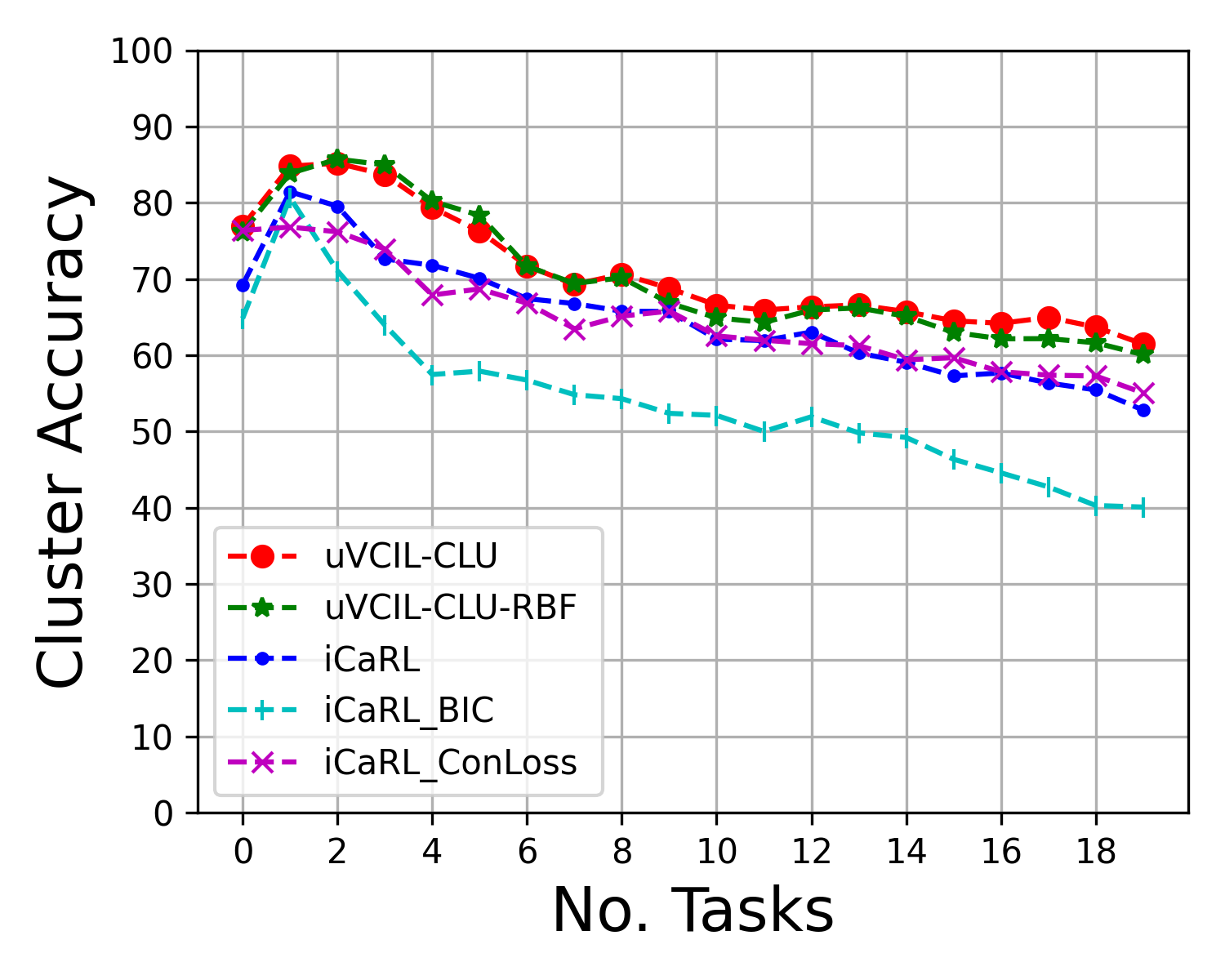}   
            \label{fig:ucf101-kmeans-acc}
            \vspace{-0.3cm}
     }
     \subfloat[HMDB51 Cluster Accuracy]{
            \includegraphics[width=0.32\textwidth]{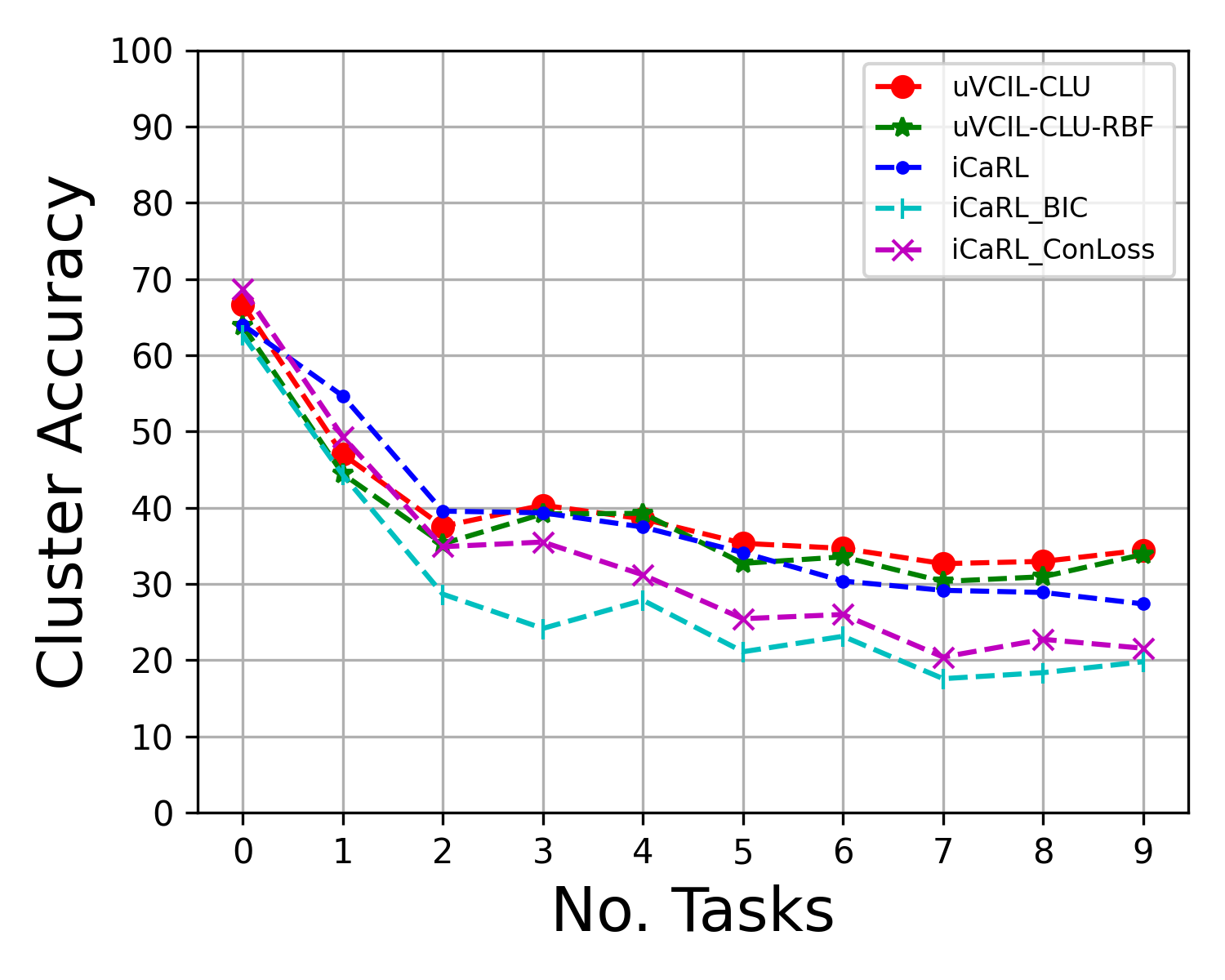}
         \label{fig:hmdb51-kmeans-acc}
         \vspace{-0.3cm}
     }
     \subfloat[SSv2 Cluster Accuracy]{
            \includegraphics[width=0.32\textwidth]{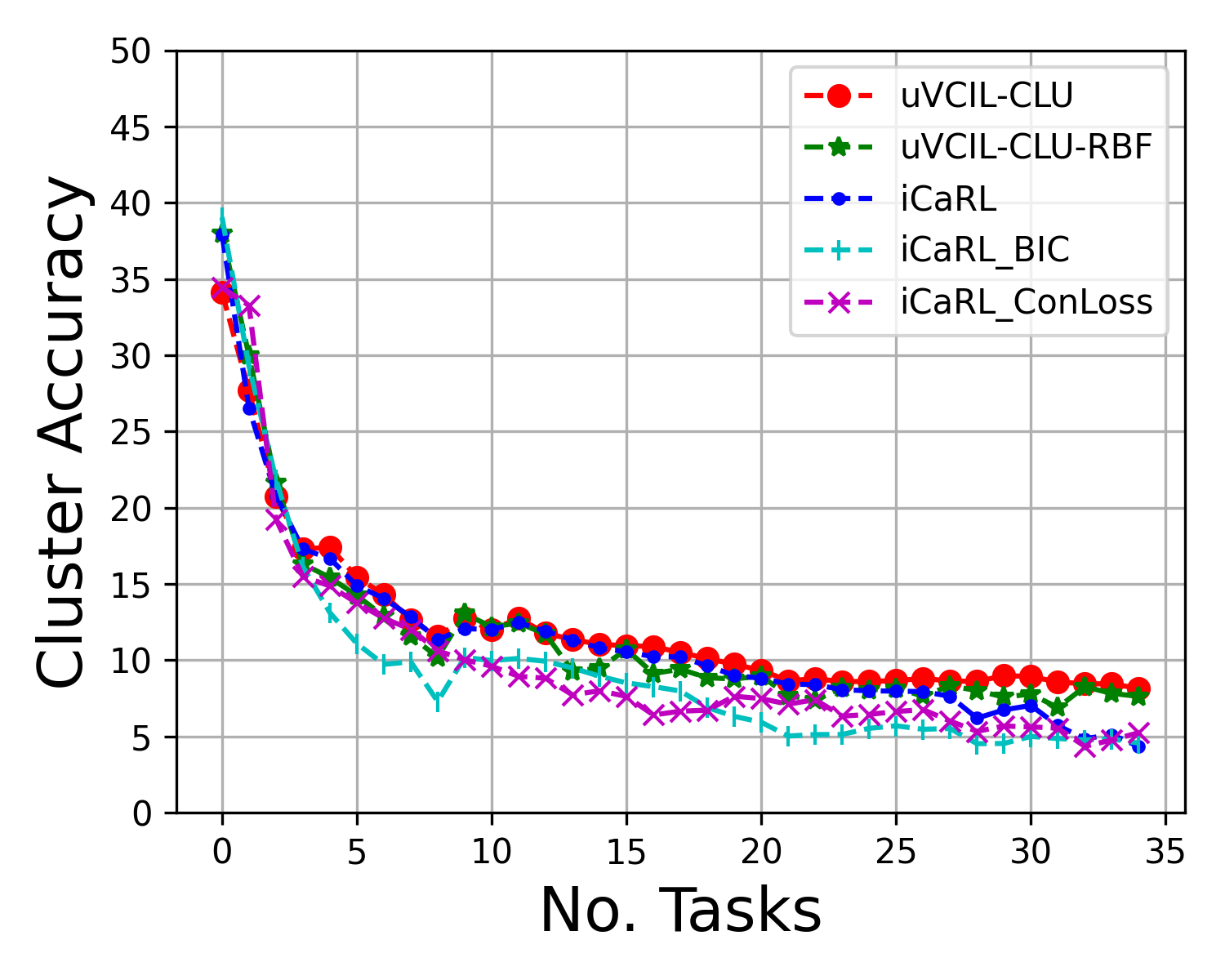}
         \label{fig:ssv2-kmeans-acc}
         \vspace{-0.3cm}
     }\vspace{-0.2cm}
     \hfill
     \vspace*{-0.1cm}
        \caption{uVCIL results on UCF101, HMDB51 and SSv2 for the first fold data based on ResNet-34 feature extractor.}
        \label{fig:main_result_learn_step}
        \vspace*{-0.5cm}
\end{figure*}
\vspace{-0.3cm}
\subsection{Linear cluster self-supervised learning}
\vspace{-0.2cm}
We proposed to employ the a Radial Basis Functions (RBF) network \cite{RBF} as linear classification layer.
The RBF network will learn the pseudo-labels $\tilde{y}_{k,j}$ allocated to the clusters, defined by the centers $\mbox{\boldmath$\mu$}_{k,j}$ at each task $\tau_k$, resulting in $L_K$ such units, while a multi-class cross entropy loss is applied to learn the pseudo-labels for the data ${\bf x}_{k,i}$. The linear part of the RBF classifier is randomly initialized and then trained with the extracted features associated with various clusters. This method is called uVCIL-CLU-RBF. This approach requires access to the data from the memory, where a memory buffer ${\cal M}_i$ is defined for each cluster, $i=1,\ldots,l_k$ and a set of $N$ data is memorized for each cluster $\{ {\bf x}_{m,i} \in {\cal M}_m\}_{i=1}^N$ for $m=1,\ldots,L_k$, where $L_k$ corresponds to the total number of clusters after learning $k$-th task, with a total of $k \times l_k \times N$ memorized data. Besides the data associated with each new task, we consider all data from existing memory clusters ${\bf M}_{k} = {\cal M}_1 \cup {\cal M}_2 \cup \ldots \cup {\cal M}_{L_k}$, where ${\bf M}_k$ represents all data stored in the memory buffers after learning tasks  $\tau_1,\ldots, \tau_k$.
Then, new pseudo-labels $\tilde{y}_{k,i}$ are allocated to the data associated with new clusters. New memory clusters are assigned to the new clusters augmenting the overall memory as ${\bf M}_{k+1} = {\bf M}_k \cup {\cal M}_{L_k+1} \cup \ldots \cup {\cal M}_{L_k+l_k} $ adding $N$ samples to each new memory buffer. Instead of storing real data in each memory buffer, we store a features vector for each video, which represents a significant reduction in the memory requirements, \cite{10647854}. This process is repeated for all tasks $i=1,\ldots,K$ leading to forming a memory ${\bf M}_K$ for combating catastrophic forgetting. When learning the new  task $\tau_{k+1}$, the weights learn at task $\tau_k$ for the $L_k$ output linear units are transferred for initializing the new classifier for the subsequent task $\tau_{k+1}$, after augmenting the number of linear units $L_k + l_{k+1}$. To address category data imbalance in uVCIL, we employ the Focal Loss (FL) \cite{FOCALLOSS}, where we replace the class concept with that of pseudo-labeled clusters. We adopt the FL for balancing the cluster weights to handle data imbalance. During training we first use the multi-class cross-entropy (MCE) loss for the Focal Loss (FL) in order to manage cluster imbalance as~:
\vspace{-0.2cm}
\begin{equation}
  FL(MCE) = \alpha_j * (1 - \exp{(-MCE)} )^\gamma *MCE,
  \label{eq:class_weight_facal_loss}
  \vspace{-0.2cm}
\end{equation}

where $\gamma=2$ is a modulating factor to multi-class cross-entropy loss and $\alpha_j$ is the pseudo-label balance weight. Our consistency regularization strategy is model-agnostic and can be adapted for imbalanced unsupervised clustering in continual learning scenarios. 

\begin{figure*}[t!]
     \centering
     \subfloat[UCF101 Task-20]{
            \includegraphics[width=0.32\textwidth]{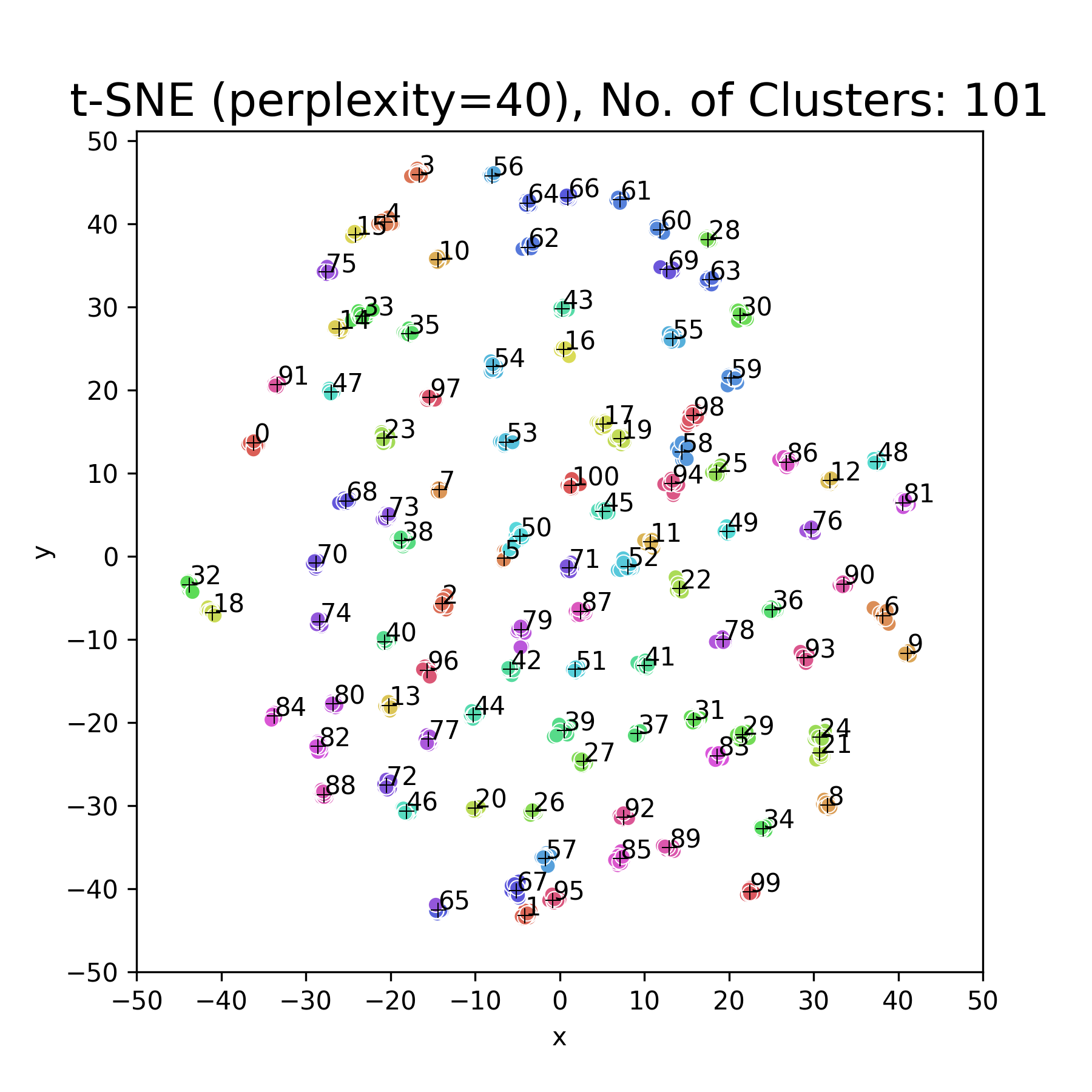}   
            \label{fig:ucf101-task20}
            \vspace{-0.3cm}
     }
     \subfloat[HMDB51 Task-10]{
            \includegraphics[width=0.32\textwidth]{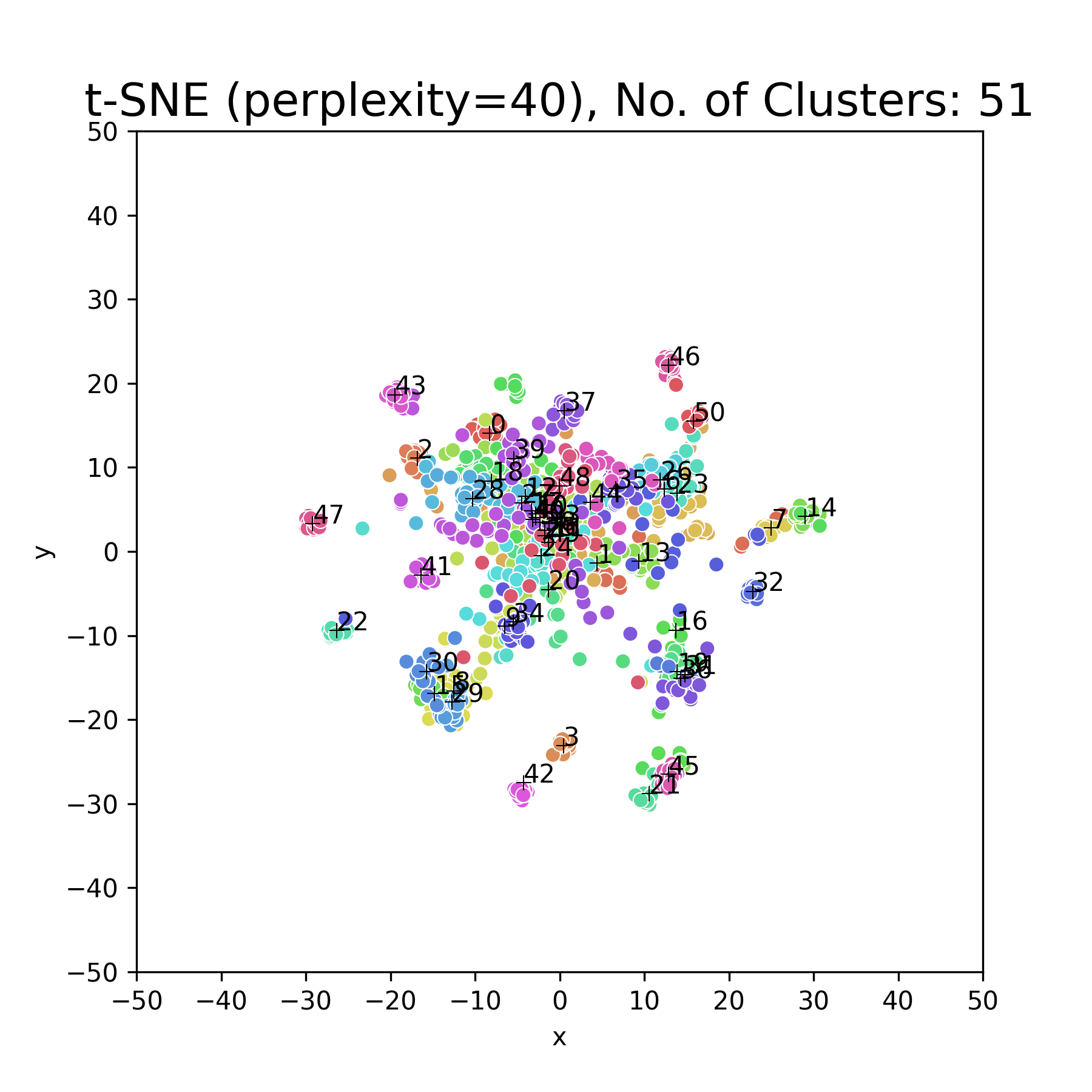}
         \label{fig:hmdb51-task10}
         \vspace{-0.3cm}
     }
     \subfloat[SSv2 Task-35]{
            \includegraphics[width=0.32\textwidth]{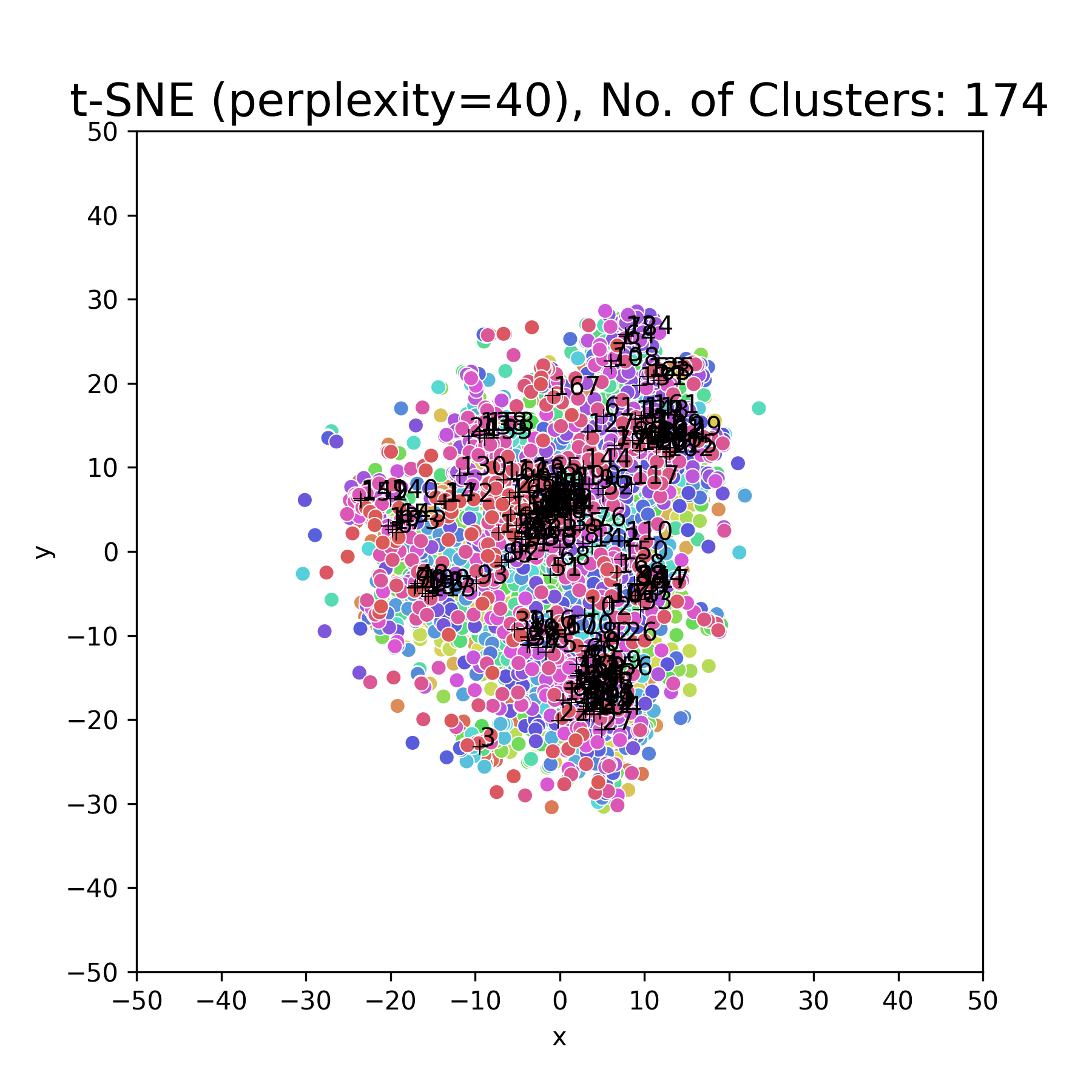}
         \label{fig:ssv2-task35}
         \vspace{-0.3cm}
     }\vspace{-0.2cm}
     \hfill
     \vspace*{-0.1cm}
        \caption{The visualization of cluster center along with their memory buffer after all subsequence learning task. This figure is best viewed in colour, $+$ represents the cluster centre, and a number represents the cluster ID. We visualize by using t-SNE for feature reduction to 2-Dimensions with a perplexity of 40.}
        \label{fig:cluster_distribution}
        \vspace*{-0.5cm}
\end{figure*}

\begin{table}[!t]
\begin{center}
\caption{uVCIL performance, on SSv2, UCF101 and HMDB51 datasets based on the ResNet-34 feature extractor. Both UCF101 and HMDB51 datasets are average of three splits.}
\vspace*{-0.2cm}
\label{tab:sumary_uvcil}
\resizebox{\linewidth}{!}{
\begin{tabular}{l|ccc|ccc|ccc}
\hline
\multirow{3}{*}{Methods}       &  \multicolumn{3}{c|}{UCF101} & \multicolumn{3}{c|}{HMDB51} &  \multicolumn{3}{c}{SSv2}         \\ \cline{2-10} 
 &    \makecell{ Avg \\ ACAcc} &   \makecell{ BWF $\uparrow$ } &   \makecell{ FWF $\downarrow$}    &   \makecell{ Avg \\ ACAcc} &   \makecell{ BWF $\uparrow$} &   \makecell{ FWF $\downarrow$}   &   \makecell{ Avg \\ ACAcc}  &   \makecell{ BWF $\uparrow$ } &   \makecell{ FWF $\downarrow$}      \\
\hline
\multirow{1}{*}{\makecell{ MAS } }           
              &                              5.41		 & \bf{-3.72}		  & 1.29 & 8.55	
	 & -5.48		 & \textbf{2.39} & 6.52		 & -4.46		 & 0.88   \\ \hline
              
 \multirow{1}{*}{\makecell{ EWC } }           
            &                              8.89	 & -6.23		  & 1.81 & 9.70	
	 & -7.56		 & 2.86 & 6.60		 & -4.55		 & 1.03    \\ \hline

 \multirow{1}{*}{\makecell{ iCaRL  } }           
             &                              64.30	 & -13.42	  & 1.23 & 36.80	 & -9.95	 & 3.60 & 11.37	 & -7.23	 & 0.99    \\
\hline
\multirow{1}{*}{\makecell{ iCaRL+BiC  } }           
            &                              53.56	 & -13.50	  & 1.68 & 28.23	 & -14.09	 & 5.16 & 9.45	 & -5.03	 & 1.01    \\
\hline
\multirow{1}{*}{\makecell{ iCaRL+CL  } }           
            &                              67.90	 & -12.73	  & 1.26 & 32.16	 & -11.77	 & 4.10 & 9.87	 & -4.76	 & \bf{0.86}    \\
\hline 
\hline 
\multirow{1}{*}{\makecell{ uVCIL-CLU (Ours)\\ } }           
             &                             70.27	
	    & -9.23		  & 0.79 & \bf{38.64}		 & -5.71		 & 2.87 & \bf{12.15}	
	 & -4.12		 & 0.76    \\ \hline
\multirow{1}{*}{\makecell{ uVCIL-CLU-RBF (Ours)\\ } }           
             &                             \bf{70.33}	& -9.49		  & \bf{0.81} & 36.27	& \bf{-4.66}		 & 2.76 & 11.53	
	 & \bf{-4.02}		 & 0.89    \\ \hline
\hline
\multirow{1}{*}{\makecell{ CLU (All in one)\\ } }           
           &                              50.79		 & -		  & - & 29.93
	 & -	 & - & 7.09	 & -	 & -    \\
\hline
 \multirow{1}{*}{\makecell{ CLU-RBF (All in one)\\ } }           
             &                             51.19    & -	  & - & 29.49
	 & -	 & - & 7.23	 & -	 & -    \\
\hline 

\end{tabular}
}
\end{center}
\vspace{-0.8cm}
\end{table}

\vspace{-0.3cm}
\section{Experimental Results}
\label{sec:experimeant}
\vspace{-0.2cm}
\subsection{Implementation Details}
Our study based on UCF101 \cite{UCF101}, HMDB51 \cite{HMDB51} and Something-to-Something V2 \cite{SSV2} datasets. We divide the data into 20 tasks for the learning of UCF101, 10 tasks for HMDB51 and 35 tasks for SSv2. Each task consist of 5 classes per task because this demonstrated the most challenging in supervised video class-incremental learning \cite{vCLIMB} and then we dropping the labels for unsupervised learning. We consider the ResNet-34 \cite{he2016deep} was pre-trained on the ImageNet dataset \cite{5206848} as the feature extractor $g(\cdot)$ to capture the encoder spatio-temporal features, as illustrated in Fig.~\ref{fig:proposed} for fair comparison. The input videos are composed of $8$ frames, of size $224 \times 224 \times 3$ pixels. The extracted features have 512 channels for each frame, which are then averaged across all frames to produce the video features. The clustering algorithms is based on the $k$-means from Scikit-learn \cite{scikit-learn}. We store the features for $N=20$ feature vector per cluster in memory buffers for uVCIL-CLU-RBF. We use PyTorch \cite{paszke2019pytorch} with a single NVIDIA GeForce GTX 1080 Ti 11GB GPU. For each task, the models are trained for up to 50 epochs, using a batch size of 8 videos. The model is optimized using Focal Loss with class weighting as in Eq.~\eqref{eq:class_weight_facal_loss}, and the Adam optimizer \cite{kingma2014adam} with a learning rate of 0.001. The lowest loss value will be used to identify the evaluation model.

\noindent\textbf{Baselines:} We re-implement five supervised video class-incremental learning methods are adapt for unsupervised learning as baselines including EWC \cite{EWC} and MAS \cite{aljundi2018memory}, iCaRL \cite{iCaRL} with and without consistency loss and the BiC \cite{BIC}. These adaptations from the open source code from vCILMB \cite{vCLIMB} with a ResNet-34 backbone \cite{he2016deep}. The $k$-means clustering is used to assign pseudo-labels. We preserve 20 videos per cluster as same as \cite{iCaRL,vCLIMB}. Therefore, the baseline memory buffer can save 3,480, 1,020, and 2,020 videos for Something-to-Something V2, HMDB51, and UCF101, respectively. 

\noindent\textbf{Evaluation Metrics:} We adapt the protocol used in the unsupervised settings from \cite{caron2018deep,van2020scan}, evaluating the cluster accuracy (CAcc), used for the unsupervised continual learning for images \cite{he2021unsupervised}. First, we employ the Hungarian matching algorithm \cite{kuhn1955hungarian} to associate each pseudo-label of a cluster with a ground truth label, where the video labels are considered only for testing and not for training. We also evaluate by using Forward Forgetting (FWF) \cite{lopez2017gradient}, and Backword Forgetting (BWF) \cite{vCLIMB,lopez2017gradient}.

\vspace{-0.3cm}
\subsection{uVCIL results on UCF101, HMDB51 and SSv2}
\vspace{-0.2cm}
In Table \ref{tab:sumary_uvcil}. both UCF101 and HMDB51, results are averaged across three different data splits. We consider two scenarios proposed in this study called uVCIL-CLU and uVCIL-CLU-RBF. The results shown that our proposed approach perform the best average cluster accuracy on all datasets. Whereas the BWF and FWF shown an equivalent performance. This can conclude that our approach is effective in achieving ACAcc and minimize the forgetting problem. Meanwhile, when pooling all data together (All in one) to do clustering at one time. We setup the $k$ for k-means equal to the ground truth number 101, 51, 174 for UCF101, HMDB51 and SSv2 respectively. This demonstrated the limitation of k-means clustering on large and varied datasets.
Moreover, we investigate the progressive learning of uVCIL. The results are presented in Fig \ref{fig:main_result_learn_step}. The results indicate that both uVCIL-CLU and uVCIL-CLU-RBF show a significant cluster accuracy over other methods in every task, as seen in Fig \ref{fig:ucf101-kmeans-acc}, \ref{fig:hmdb51-kmeans-acc}, and \ref{fig:ssv2-kmeans-acc} for UCF101, HMDB51, and SSv2, respectively.

\vspace{-0.3cm}
\subsection{Ablation Study}
\label{sec:ablation}
\vspace{-0.2cm}
\textbf{Supervised vs Unsupervised feature extractor:}
To avoid bias from the supervised pre-trained scenario. We are considering two different pre-training strategies. First, ResNet-34 \cite{he2016deep} are considered as the supervised pre-trained model on the ImageNet dataset. Second, VideoMAE-v2 \cite{wang2023videomae} is known as an unsupervised pre-training model, pre-trained on the Unlabeld Hybrid-1M dataset without any ground truth labels. Each video input contains 16 frames. The results provided in Table~\ref{tab:ucf101_hmdb51_change_backbone}, it is confirmed that our proposed approach performs well on both the supervised and unsupervised feature extractors expecially on unsupervised video feature extractor.

\begin{table}[ht!]
\begin{center}
\caption{uVCIL performance on different feature extractors: VideoMAEv2 unsupervised (U), and ResNet-34 supervised (S) pre-trained scenarios. }
\label{tab:ucf101_hmdb51_change_backbone}
\vspace{-0.3cm}
\resizebox{\linewidth}{!}{
\begin{tabular}{lc|ccc|ccc|ccc}
\hline
 \multirow{ 3}{*}{\makecell{Methods}}   & \multirow{ 3}{*}{\makecell{ Feature \\Extractor\\ }}    &  \multicolumn{3}{c|}{UCF101} & \multicolumn{3}{c|}{HMDB51}   & \multicolumn{3}{c}{SSv2}       \\ \cline{3-11}   
    &       &  \makecell{  Avg \\ACAcc} &   \makecell{ BWF $\uparrow$ } &   \makecell{ FWF $\downarrow$}  & \makecell{  Avg \\ ACAcc} &   \makecell{ BWF $\uparrow$ } &   \makecell{ FWF $\downarrow$}  & \makecell{  Avg \\ACAcc} &   \makecell{ BWF $\uparrow$ } &   \makecell{ FWF $\downarrow$}        \\
\hline
\multirow{2}{*}{\makecell{uVCIL-CLU}}
& S                 &            68.36	
	    & -7.88		  & 0.39 & 37.74	 & -4.94		 & 2.80 & 12.46	
	 & -4.40		 & \bf{0.80}   \\
                  &   U               &            \bf{98.33}  &                 \bf{0.01}	 & \bf{0.09}  & \bf{62.84}	 & -1.79	 & 2.53  & \bf{15.62}	 & -4.40	 & 1.06 \\ \hline
\multirow{1}{*}{\makecell{ CLU (All in one)\\ } }           
            & U   &                             95.97	    & -	  & - & 57.73	 & -	 & - & 8.41 	 & -	 & -    \\ \hline \hline
 \multirow{2}{*}{\makecell{uVCIL-CLU-RBF}}
                  & S                 &           68.88	
	    & -7.69	  & 0.33 & 36.64	 & -7.01		 & 2.96 & 11.61	
	 & \bf{-4.73}		 & 0.94 \\
                  & U             &            \bf{97.40} &  \bf{-3.40}	 & \bf{0.31}  & \bf{60.49}	 & -4.17	 & 3.31 & \bf{15.40}	 & -4.97	 & 1.08  \\
                 \hline
                 
\multirow{1}{*}{\makecell{ CLU-RBF (All in one)\\ } }           
            & U  &                             95.97	    & -	  & - & 57.50	 & -	 & - & 8.69	 & -	 & -    \\ 
\hline 

\end{tabular}
}
\vspace{-0.7cm}
\end{center}
\end{table}

\noindent\textbf{The cluster visualization:} We visualiz the progressive cluster of uVCIL-CLU-RBF with VideoMAEv2 feature extractor are presented in Fig \ref{fig:cluster_distribution}. The results indicate that on UCF101 and HMDB51 are well separately cluster, whereas on the SSv2 with is the most challenging dataset, which is plenty room to improving in the future study.

\noindent\textbf{Computation cost:} For UCF101 dataset, our uVCIL-CLU use 51.71K trainable parameter and uVCIL-CLU-RBF use 52.83K trainable parameter. Whereas the baseline methods use 21.3M trainable parameter. This results shows that our proposed approach can learn roughly 199\% faster than baseline approaches. Because our approach does not require fine tune the feature extraction while the baseline are require.

\vspace{-0.3cm}
\section{Conclusion}
\label{sec:conclusion}
\vspace{-0.2cm}

In this paper, we propose the Unsupervised Video Class-Incremental Learning (uVCIL), which introduces a hierarchical, expanding memory system that relies on the continuously extraction of visual features from video data. Then it builds a deep embedded clustering network using the video features. We propose two different approaches, based on the $k$-means pseudo-label guide for classification and finding relevant clusters. The key to avoiding forgetting is that of managing memory buffers, storing features of videos associated with each cluster, which are afterwards used for memory replay. When new tasks are introduced for learning, the model recalls prior information through memory replay in order to mitigate catastrophic forgetting. Our experiments highlight that proposed approach effectively preserves past knowledge while learning computationally efficient new information. For future work we will focus on improving feature representation to tackle domain shift problem for uVCIL .

\vfill\pagebreak

\bibliographystyle{IEEEbib}

\scriptsize{
\bibliography{strings}
}
\end{document}